\documentclass[conference]{IEEEtran}
\IEEEoverridecommandlockouts
\usepackage{cite}
\usepackage{amsmath,amssymb,amsfonts}
\usepackage{algorithmic}
\usepackage{graphicx}
\usepackage{textcomp}
\usepackage{xcolor}
\usepackage{hyperref} 

\def\BibTeX{{\rm B\kern-.05em{\sc i\kern-.025em b}\kern-.08em
    T\kern-.1667em\lower.7ex\hbox{E}\kern-.125emX}}
\begin{document}

\title{Prediction of Stocks Index Price using Quantum GANs}

\author{
    Sangram Deshpande\textsuperscript{1, 2}, 
    Gopal Ramesh Dahale\textsuperscript{1}
    Sai Nandan Morapakula\textsuperscript{1, 3}, 
    Dr. Uday Wad\textsuperscript{1},
    \\
    \\

    \textsuperscript{1} Qkrishi Quantum Pvt. Ltd., Gurgaon, Haryana, India \\
    \textsuperscript{2} Electrical and Computer Engineering, NC State University, Raleigh, USA \\
    \textsuperscript{3} Physics Department, University of Massachusetts, Boston, MA, USA \\
    Corresponding author: Sangram Deshpande email: ssdesh24@ncsu.edu 
}

\maketitle

\begin{abstract}
 This paper investigates the application of Quantum Generative Adversarial Networks (QGANs) for stock price prediction. Financial markets are inherently complex, marked by high volatility and intricate patterns that traditional models often fail to capture. QGANs, leveraging the power of quantum computing, offer a novel approach by combining the strengths of generative models with quantum machine learning techniques. We implement a QGAN model tailored for stock price prediction and evaluate its performance using historical stock market data. Our results demonstrate that QGANs can generate synthetic data closely resembling actual market behavior, leading to enhanced prediction accuracy. The experiment was conducted using the Stocks index price data and the AWS Braket SV1 simulator for training the QGAN circuits. The quantum-enhanced model outperforms classical Long Short-Term Memory (LSTM) and GAN models in terms of convergence speed and prediction accuracy. This research represents a key step toward integrating quantum computing in financial forecasting, offering potential advantages in speed and precision over traditional methods. The findings suggest important implications for traders, financial analysts, and researchers seeking advanced tools for market analysis.

\end{abstract}




\section{{Introduction}}

Accurate price prediction can aid in determining risk exposure, setting margin limits, and issuing margin calls, among other things. Nevertheless, the volatile nature of markets and the influence of multiple factors, such as policy changes, interest rate shifts, and currency fluctuations, make this process complex. Stock price prediction is essential for investors, financial analysts, and traders as it provides insights into future market trends. Accurate predictions enable investors to make informed decisions regarding buying, holding, or selling stocks, which can significantly affect their financial returns. Given the inherent volatility and complexity of financial markets, effective prediction models are crucial for risk management, portfolio optimization, and strategic planning. Reliable stock price forecasts help in minimizing losses and maximizing profits by identifying potential opportunities and threats in the market. For individual investors, it means better investment strategies and higher returns. For financial institutions, it enables the development of sophisticated trading algorithms, enhancing the efficiency and profitability of their trading operations. Companies can use these predictions to make strategic business decisions, such as timing for issuing new shares or buybacks. Additionally, accurate forecasts contribute to market stability by reducing the likelihood of large, unexpected price swings. This predictability can also foster investor confidence, attracting more capital into the markets and supporting overall economic growth. \cite{1}

Fully Quantum Generative Adversarial Networks (QGANs) are advanced machine learning models that leverage the principles of quantum computing to enhance generative modeling capabilities. Traditional GANs consist of a generator and a discriminator, where the generator creates synthetic data, and the discriminator evaluates its authenticity.\cite{2} QGANs incorporate quantum algorithms into this framework, potentially providing exponential speed-ups and improved accuracy due to quantum parallelism and entanglement. In stock price prediction, QGANs can process complex patterns and correlations in historical market data more efficiently than classical models, generating realistic synthetic data that closely mirrors actual market behaviors. This capability allows QGANs to produce more accurate and reliable stock price forecasts, aiding in better decision-making for investors and financial analysts. \cite{30}

Classical Generative Adversarial Networks (GANs) in finance are primarily used for generating synthetic financial time-series data, which is valuable for backtesting trading strategies, risk assessment, and stress testing financial models. They can capture complex patterns in financial data, such as volatility and correlations. Additionally, GANs are used for anomaly detection in financial data, identifying potential fraud, market manipulation, or sudden price changes. By learning normal patterns, they can pinpoint suspicious deviations. Classical GANs are also employed to enhance financial forecasting by creating more realistic data for training predictive models, thereby improving the accuracy of stock price forecasts. They aid in risk management by generating various market scenarios, helping financial institutions better understand their risk exposure.

\section{Problem formulation: GANs to QGANs}

\begin{figure}
  \centering
  \includegraphics[width=0.5\textwidth]{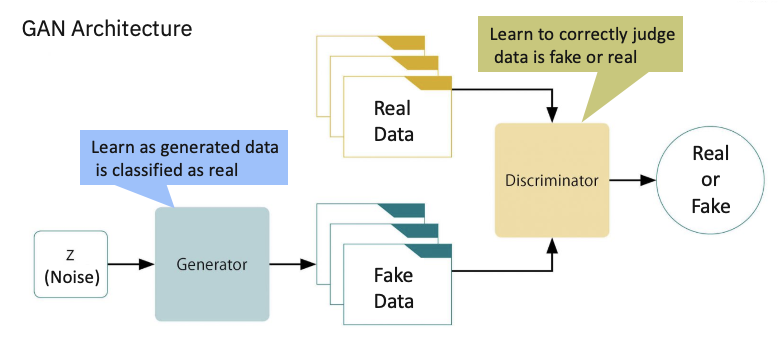} 
  \caption{GAN Architecture from paper \cite{29}}
  \label{figure1}
\end{figure}

We propose a generalized Quantum Generative Adversarial Network that takes advantage of the classical GAN architecture integrated with quantum-enhanced prediction power. \cite{25} GANs are a popular artificial intelligence model that consists of two neural networks, as shown in Figure \ref{figure1}:
\begin{itemize}
    \item Generator: Replicates a real-world dataset by producing synthetic data samples (e.g., text, audio, and/or photos).
    \item Discriminator: Assists in differentiating between authentic data and the generator's synthetic examples.
\end{itemize}

To ``fake`` realistic historical and future stock price data, mimicking real market dynamics and capturing complex relationships, the two competing entities, Generator and Discriminator, collaborate when using GANs for stock price prediction. The other part of the system acts as a critic, attempting to discern fake data from the real historical price data. As a result of their ongoing competition, the discriminator improves at spotting departures from actual market trends, while the generator gradually grows more skilled at producing realistic and representative data. The idea behind this is that after going through this training process, the generator could be able to provide ``future'' market data that almost exactly matches current prices, which could aid in prediction.

\begin{figure}
  \centering
  \includegraphics[width=0.5 \textwidth]{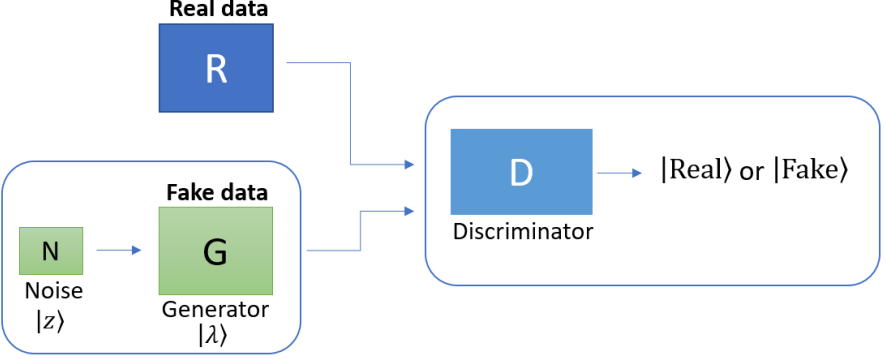} 
  \caption{QGAN Architecture from paper \cite{28}}
  \label{}
\end{figure}

\vspace{10mm}
\textbf{Training Loop and Feedback}
\begin{itemize}
    \item The generator is trained based on input from the discriminator, which pushes it to produce more realistic data that can trick the discriminator. The generator and discriminator are simultaneously improved by this iterative adversarial process. \cite{5}
    \item The model's performance is determined by calculating the difference between the discriminator's predicted probability and the true label (real or created) using the loss function.
    \item The discriminator's weights and biases are updated using the loss using backpropagation, a method of modifying model parameters in response to prediction mistakes. The goal of this procedure is to increase the discriminator's capacity to differentiate between produced and actual data in later rounds.
\end{itemize}

The iterative adversarial process improves both the generator and discriminator simultaneously.
Using a loss function, model performance is determined by evaluating the difference between predicted probabilities and true labels. Backpropagation updates the discriminator's weights and biases, enhancing its ability to differentiate between produced and actual data in subsequent rounds.

In the realm of QGANs, the discriminator employs a supervised learning technique, often a Quantum Deep Neural Network (DNN). \cite{20}
Maintaining a delicate balance between the discriminator's ability to identify fakes and the generator's capacity to produce realistic data ensures continuous improvement throughout training.

Like in classical techniques, to categorise data as genuine or produced, the discriminator in QGAN usually uses a supervised learning technique, most often a quantum deep neural network (DNN). Unstable QGAN training and less-than-ideal outcomes might arise from a discriminator with inadequate training. The discriminator's capacity to identify fakes and the generator's capacity to generate realistic data must be balanced. This equilibrium makes sure that throughout training, both components become better repeatedly.

\section {Data preparation} 

Data preparation is an essential initial phase in any predictive modeling endeavor, including those employing Quantum Generative Adversarial Networks (QGANs) for stock price prediction. This process encompasses several critical steps, beginning with the acquisition of pertinent historical financial data, stock prices, trading volumes, and economic indicators, sourced from reputable databases or market exchanges. Subsequent preprocessing involves cleaning and transforming raw data, handling missing values, outliers, and normalizing or scaling to ensure consistency and comparability. Feature engineering follows, creating new variables or features to capture relevant patterns within the data, such as technical indicators or additional market-related factors.

We have used a real-market dataset for our analysis, specifically focusing on the FTSE stock price index. The dataset spans 10 years.\cite{8}

Another vital aspect is data encoding, converting processed data into a format compatible with the QGAN architecture, typically involving numerical vectorization or tensor conversion. Finally, data partitioning divides the dataset into training, testing sets to facilitate model training, evaluation, and validation. This partitioning ensures that the model's performance is assessed on unseen data, guarding against overfitting and enhancing generalization to new market conditions. In summary, meticulous data preparation establishes the groundwork for developing robust predictive models, enabling the effective utilization of QGANs for stock price prediction.

\section{QGANs}
Quantum Generative Adversarial Networks (QGANs) represent a novel approach within the realm of quantum computing, mirroring their classical counterparts but operating within the quantum framework. As described in textbooks, QGANs consist of two main components: the qGenerator and the qDiscriminator. The qGenerator is tasked with producing synthetic data samples, while the qDiscriminator distinguishes between real and generated data. 

Training a QGAN involves an iterative process where both components are trained concurrently, fostering a competitive dynamic. This adversarial training encourages the qGenerator to produce increasingly realistic data, while the qDiscriminator improves its ability to discern between real and synthetic samples. \cite{7, 32, 31}

Real-world market data may closely resemble synthetic financial data produced by QGANs, despite the inherent idiosyncrasies and complexity. This makes it feasible to get beyond limitations on real data availability and variety, which enhances training datasets for prediction models. By generating a range of scenarios and market swings, QGANs may help assess how robust prediction models are in several contexts, leading to more adaptable and durable forecasts. By identifying hidden patterns, QGANs may be able to identify components that conventional models would miss by using the entanglement characteristics of qubits\cite{28}. Breakdown of QGAN's operational parts:
\begin{itemize}
        \item \textbf{qGenerator:}
            \begin{itemize}
                \item \underline{Data Encoding}: Historical prices, volume, and economic indicators of real-world stocks are pre-processed and encoded into vectors that are compatible with the GAN architecture.
                \item \underline{Noise Injection}: Adding random noise to the encoded data is known as ``noise injection". In addition to adding variation, this keeps the generator from just memorising past data.
                \item\underline{Quantum Operations}: Quantum circuits and operations may be employed in more sophisticated methods to represent intricate non-linear correlations seen in the financial data, possibly producing more realistic simulations.
                \item \underline{Deep Learning Network}: After being processed by several deep learning layers (such as convolutional or recurrent neural networks), the noisy encoded data is refined to produce the ``fake" pricing data sequence.
            \end{itemize}
        \item \textbf{qDiscriminator:}
            \begin{itemize}
                \item \underline{Feature Extraction}: Similar to the generator, the discriminator uses deep learning layers to extract features from both data sets.
                \item \underline{Data Comparison}: The discriminator receives both the real historical price data and the generated data from the generator.
                \item \underline{Real/Fake Classification}: Based on the extracted features, the discriminator outputs a probability score indicating its confidence in whether the data is real or generated.
            \end{itemize}
\end{itemize}
\begin{figure}[h!]
  \centering
  \includegraphics[width=0.3\textwidth]{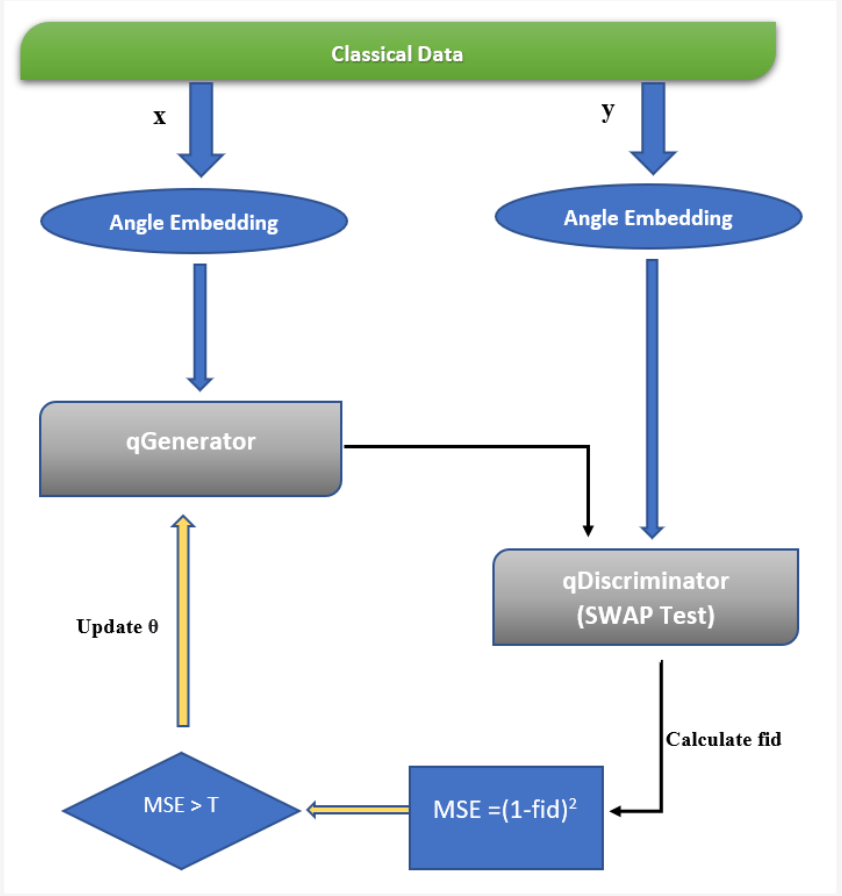} 
  \caption{Proposed QGAN Model for Stock Price Prediction}
  \label{}
\end{figure}

One of the notable advantages of QGANs lies in their capacity to capture intricate patterns and relationships within datasets, including those with quantum characteristics. This feature is particularly advantageous in domains such as stock price prediction, where conventional models may struggle to capture the complexities of financial markets.

In essence, QGANs offer a promising avenue for data generation and manipulation within the quantum computing domain, with potential applications spanning various fields, including finance, healthcare, and materials science.

\subsection{Hybrid Quantum GAN}

\begin{figure}[h!]
    \centering
    \includegraphics[width=0.6\linewidth]{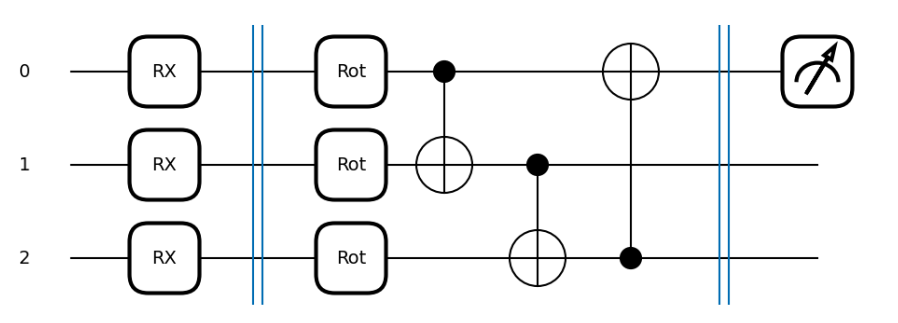}
    \caption{Quantum circuit as a generator for Hybrid QGAN for a past window of 3 days and future of 1 day. Pauli-Z observable on the first qubit.}
    \label{fig:hybrid gen}
\end{figure}

In the hybrid quantum GAN, we replace the classical generator with the quantum circuit as shown in Fig.~\ref{fig:hybrid gen}. The discriminator remains classical. The data is encoded as angles of rotation gates. The number of qubits grows with the size of the past window.

\subsection{Fully Quantum GAN}
We have managed to implement a fully Quantum GAN during our phase 2 testing. This was done by removing the conventional classical discriminator and replacing it with a swap test.  
At its core, our QGAN architecture comprises three key components:
\begin{itemize}
    \item \underline{Data Encoder}: This component plays a pivotal role in the transformation of classical input data into quantum states. By embedding real-world financial data into quantum states, the Data Encoder initiates the quantum computing process and facilitates subsequent quantum operations.
    \item \underline{Variational Quantum Circuit (VQC) based Generator}: The VQC-based Generator constitutes the heart of the QGAN architecture, responsible for generating synthetic data that closely mimics real-world market dynamics. Leveraging the principles of variational quantum circuits, this component employs quantum operations to generate synthetic data sequences with high fidelity and realism.
    \item \underline{SWAP Test-Based Discriminator}: In a departure from traditional discriminator architectures, the SWAP Test-based Discriminator represents a fully quantum approach to discerning between real and synthetic data. By employing SWAP tests, which measure the fidelity between quantum states, this discriminator evaluates the similarity between the real and fake data generated by the VQC-based Generator. Also, there are no parameters that need to be trained here. 
    \item \underline{Window Size and Prediction Horizon}: The QGAN architecture adopts a specific window size and prediction horizon, with a 3-day window utilized for predicting 1-day stock price movements. To encode relevant features of the financial data, angle encoding is employed, chosen for its noise immunity and simplicity of implementation within the quantum framework. 
\end{itemize}


\subsubsection{Architecture for larger future window}

To incorporate a larger future window in the FQGAN(Fully Quantum GAN), we used Amplitude Embedding to encode data in the quantum circuit. We provide an overview of the pipeline used:

\begin{enumerate}
    \item The adjusted closing price data is used. We eliminate the noise in the data by applying a Hodrick-Prescott \cite{34} smoothing, which extracts the data, followed by splitting the data into training (80\%) and testing (20\%) datasets.
    \item We perform min-max scaling to scale the prices between 0 and 1, followed by normalizing the datasets, i.e l2 norm should be 1. The normalization is essential for amplitude embedding.
    \item Train the FQGAN. Used the trained weights to obtain predictions.
    \item Inverse the normalization and min-max scaling to obtain the predicted future prices. Reapply the Hodrick-Prescott to eliminate the noise in predictions.
\end{enumerate}

\begin{figure}[h!]
\centering
   \begin{minipage}{0.49\textwidth}
     \centering
     \includegraphics[width=\linewidth]{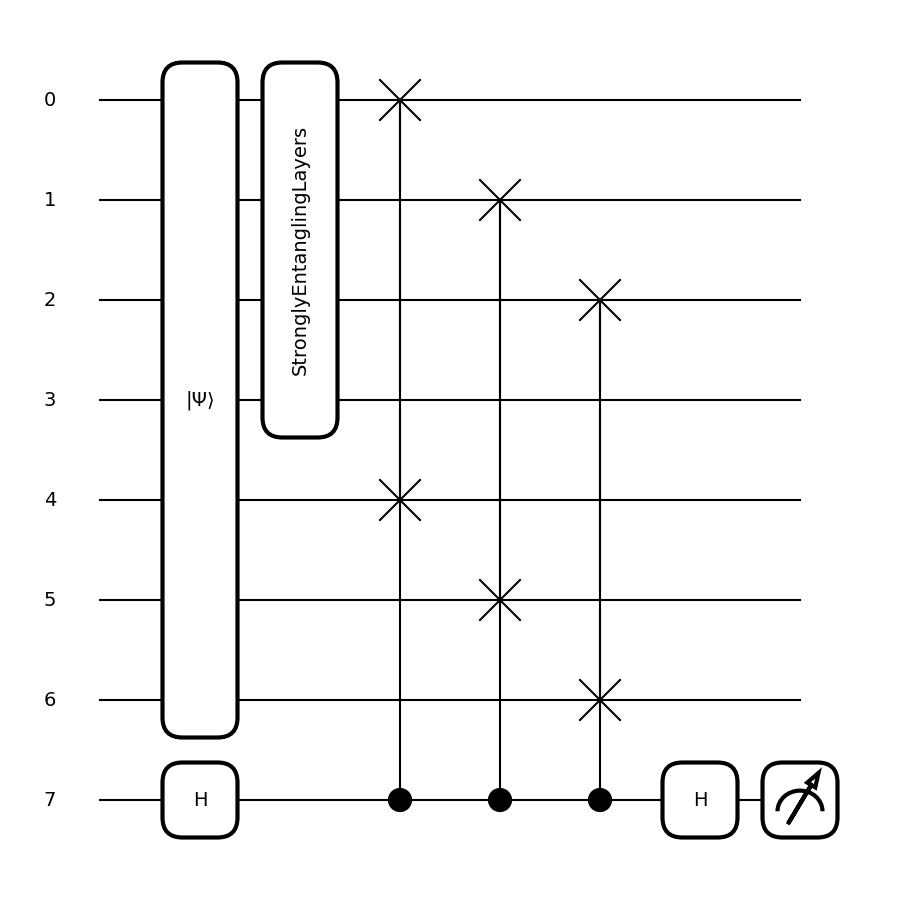}
   \end{minipage}\hfill
\end{figure}

The number of qubits scales based on the following relation

\begin{equation}\label{eq:qubit}
    \lceil \log_2 b \rceil + \lceil \log_2 f \rceil + 1
\end{equation}

Where $b$ (backward) is the size of the past window and $f$ (forward/future) is the size of the future window. One additional qubit is required for the SWAP test measurement.

\subsubsection{Limitation}

One limitation of this architecture is that we need to store the normalization factors on the train and test datasets to perform the inverse transform. In reality, this is not available for the test dataset (as we are predicting the unknown). With the Quantum Generator, we can only obtain the normalized prices for the future window. This is a limitation of the proposed FQGAN methodology.

\subsubsection{Overcoming the Limitation}

To overcome the limitation of the normalization factor, we use a simple strategy with which we can obtain the normalization factor for the predictions from the training dataset. We refer to the FQGAN, which uses this strategy as Invertible FQGAN. We explain the strategy with the help of an example.

\begin{figure}[h!]
\centering
   \begin{minipage}{0.49\textwidth}
     \centering
     \includegraphics[width=0.6\linewidth]{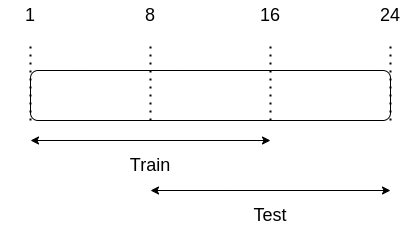}
   \end{minipage}\hfill
   \begin{minipage}{0.49\textwidth}
     \centering
     \includegraphics[width=0.8\linewidth]{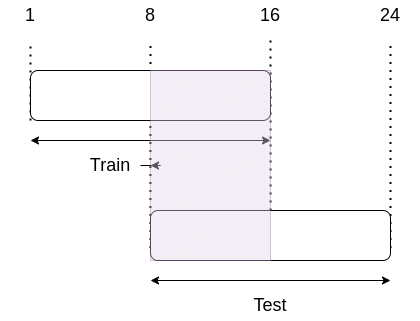}
   \end{minipage}
   \centering
   \caption{Data preparation for Invertible FQGAN. We use overlapped train and test data to train the model(left). Once trained, the Quantum Generator can generate 16 values in the future, which have an overlap with the input data and therefore the normalization factor can be obtained (right).}
   \label{fig:overlap}
\end{figure}

Consider a past window of size 16 and a future window of size 8, which means that we need to predict 8 values. Instead of predicting 8 values, we try to predict 16 values in the future, i.e. we predict the 8 unknowns as well as the 8 knowns that are in the past window of size 16. Once trained on this type of data, given a past window of size 16 as input to the Quantum Generator, we can generate 16 values in the future out of which the first 8 overlap with the input data (Fig~\ref{fig:overlap}). Using classical optimization, we try to minimize the following objective function to find the normalization factor.

\begin{equation}
    \sum_{i=1}^8 (a_i - f\hat{y_i})^2 
\end{equation}

Where $a_i$ is the min-max scaled price of the overlapping input data, $\hat{y_i}$ is the normalized predicted price of the overlapping part, and $f$ is the normalization factor. 

Once $f$ is found, we can multiply it with the $\hat{y_i}$ and then perform the inverse min-max transform to obtain the prediction prices. 

\section{Results}

The results presented in this section encompass the outcomes of our experiments, offering a comprehensive analysis of the performance of each model—Classical GAN, Hybrid Quantum-Classical GAN, and Fully Quantum GAN—based on key evaluation metrics: Root Mean Square Error (RMSE), Mean Absolute Error (MAE), and the coefficient of determination (R² score). This comparative analysis not only highlights the advantages and limitations of quantum computing in financial forecasting but also provides a foundation for future research avenues in this emerging field. By contributing to the ongoing discourse on the feasibility of quantum-enhanced predictive models, this study paves the way for significant advancements in stock market prediction.

\subsection{One Step Forecast Results}

We present results from experiments conducted with classical, hybrid, and fully quantum Generative Adversarial Networks (GANs), using a past window size of 3 days to predict the stock index for the next day. This window size strikes a balance between capturing short-term market trends and limiting computational complexity.

\subsubsection{Results with Classical GAN}

We calculated the most popular technical indicators for classical GAN, including 7-day and 21-day moving averages, exponential moving average, and momentum. Along with this, we created Fourier transforms to extract long-term and short-term trends in the stock prices. 

 \begin{figure}
  \centering
  \includegraphics[width=0.5\textwidth]{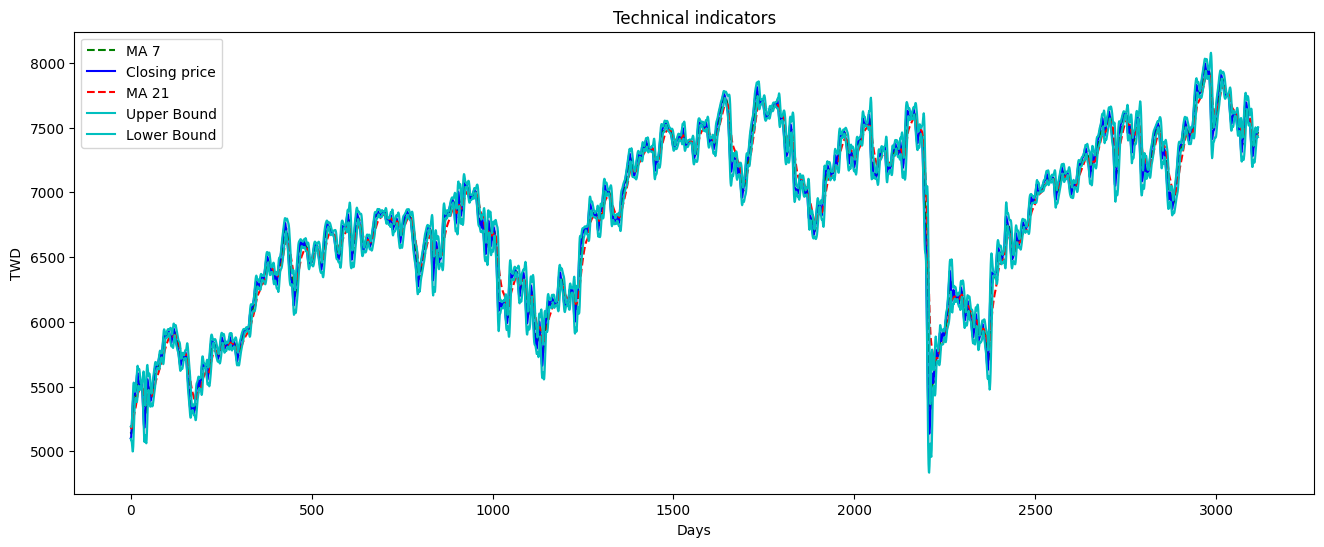} 
  \caption{Various technical indicators: Mean average, Momentum, etc. }
  \label{fig:technical_indicators}
\end{figure}

 \begin{figure}
  \centering
  \includegraphics[width=0.5 \textwidth]{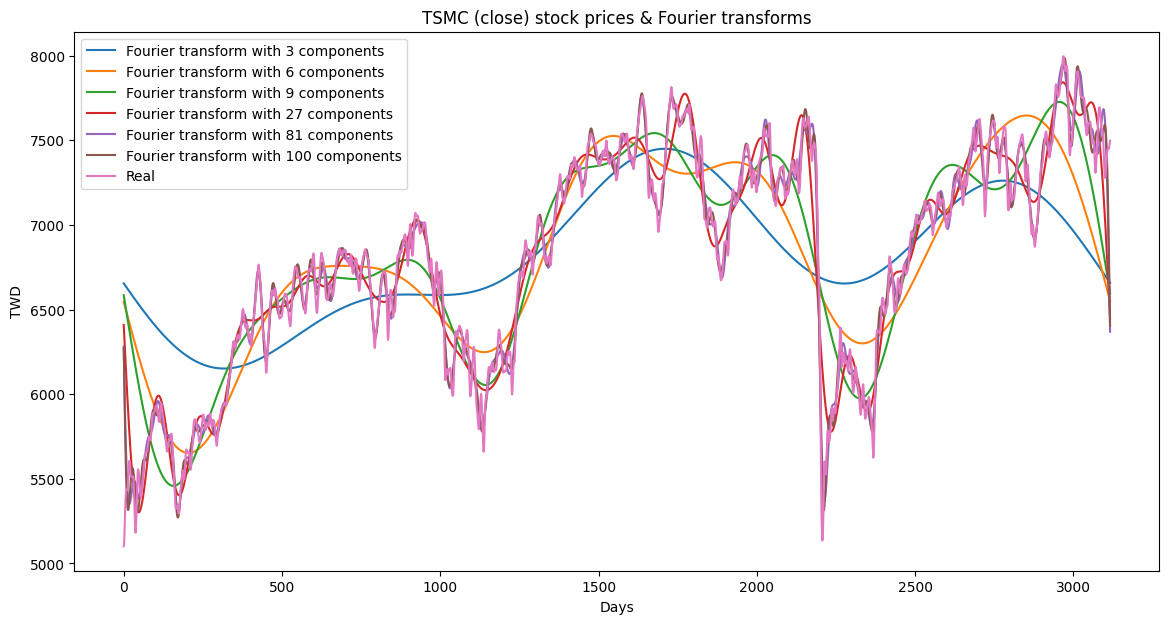} 
  \caption{Fourier transforms}
  \label{fig:fourier_transform}
\end{figure}

The Classical GAN model was used as a baseline for comparing the performance of quantum-enhanced models. The study involved two variants: one that incorporated technical indicators (referred to as Classical GAN with TI), and another that used only adjusted closing prices (referred to as Simple GAN).

\begin{figure}
\centering
   \begin{minipage}{0.49\textwidth}
     \centering
     \includegraphics[width=\linewidth]{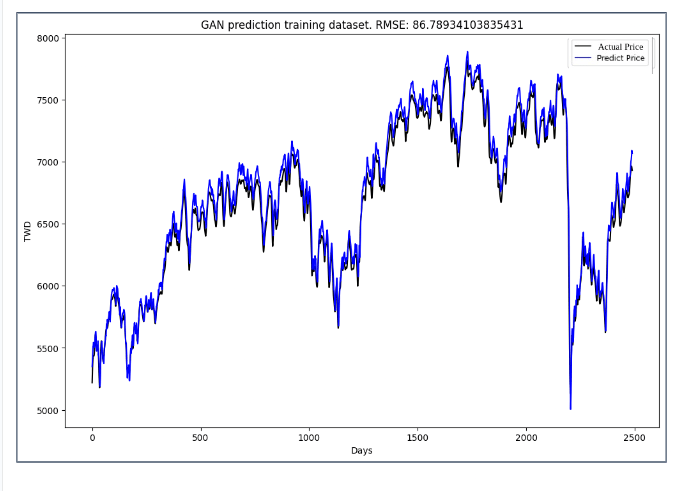}
   \end{minipage}\hfill
   \begin{minipage}{0.49\textwidth}
     \centering
     \includegraphics[width=\linewidth]{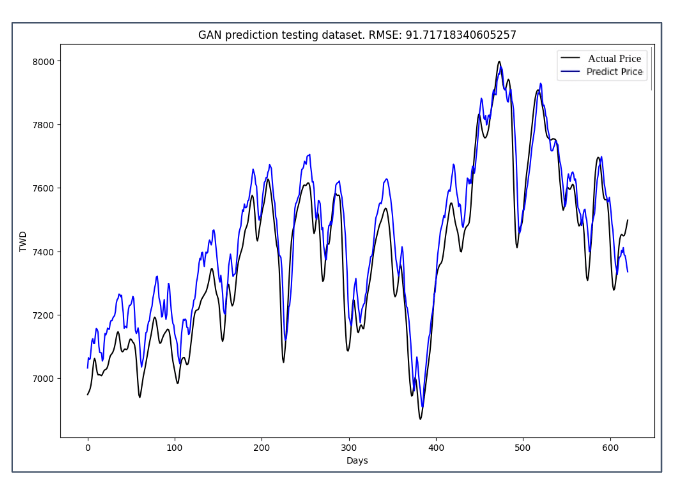}
   \end{minipage}
   \centering
   \caption{Classical GAN with Technical Indicators. Train (Up) and test (down).}
   \label{fig:classical_gan_technical_indicators}
\end{figure}

\textbf{Performance Metrics:} The Classical GAN with TI model underwent training for 150 epochs with a learning rate of 0.00016 and a batch size of 128, utilizing the Adam optimizer. The performance was primarily evaluated using RMSE, providing a quantitative measure of the deviation between predicted and actual stock prices. Training and test performance results are presented in Fig.~\ref{fig:classical_gan_technical_indicators}.

It should be noted that these classical results are taken as standards, and we compare our other method results with these.

\subsubsection{Results with Hybrid Quantum GAN}
The Hybrid Quantum-Classical GAN marks an important step in integrating quantum computation with classical models. In this approach, the generator employs a quantum circuit while the discriminator is based on classical architecture. Despite similar hyperparameters to the classical model, the Hybrid Quantum GAN demonstrated distinct performance characteristics.

\begin{figure}
\centering
   \begin{minipage}{0.49\textwidth}
     \centering
     \includegraphics[width=\linewidth]{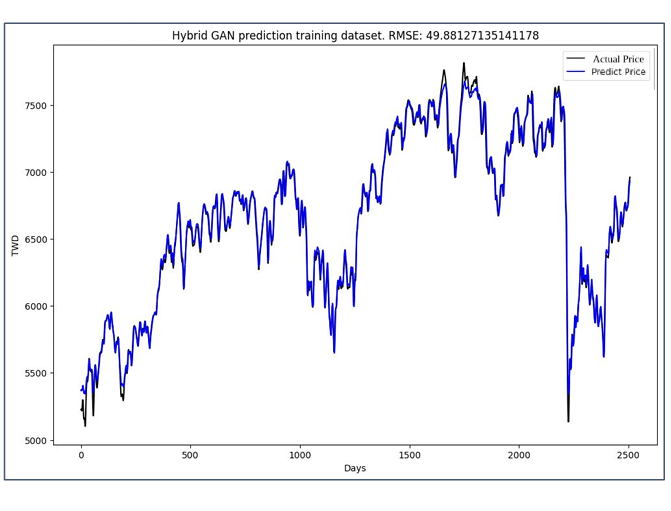}
   \end{minipage}\hfill
   \begin{minipage}{0.49\textwidth}
     \centering
     \includegraphics[width=\linewidth]{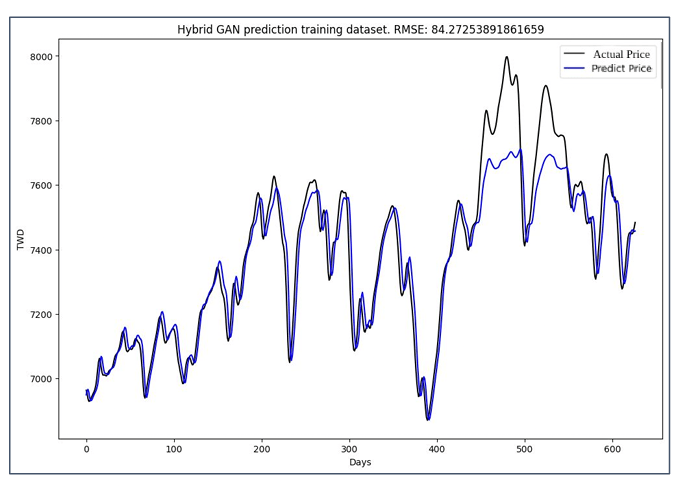}
   \end{minipage}
   \centering
   \caption{Hybrid Quantum GAN. Train (Up) and test (Down).}
   \label{fig:hybrid_qgan_trainNtest}
\end{figure}

\textbf{Performance Metrics:} Trained over 150 epochs with the same learning rate and batch size as the Classical GAN, the Hybrid Quantum GAN achieved a training RMSE of 49.88 and a higher test RMSE of 84.27, indicating that the quantum component introduced significant variability in model performance. These results highlight the challenges associated with combining quantum and classical systems in stock price prediction. Fig. \ref{fig:hybrid_qgan_trainNtest}

\subsubsection{Results with Full Quantum GAN}
The Fully Quantum GAN (FQGAN) leverages both the generator and discriminator as quantum circuits. This model seeks to fully exploit quantum computing’s advantages, such as parallelism and entanglement, to capture complex stock price patterns.

\begin{figure}[h!]
\centering
   \begin{minipage}{0.49\textwidth}
     \centering
     \includegraphics[width=\linewidth]{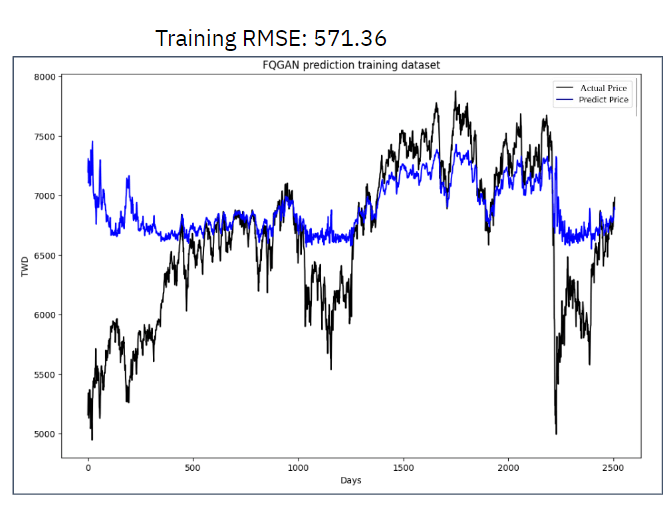}
   \end{minipage}\hfill
   \begin{minipage}{0.49\textwidth}
     \centering
     \includegraphics[width=\linewidth]{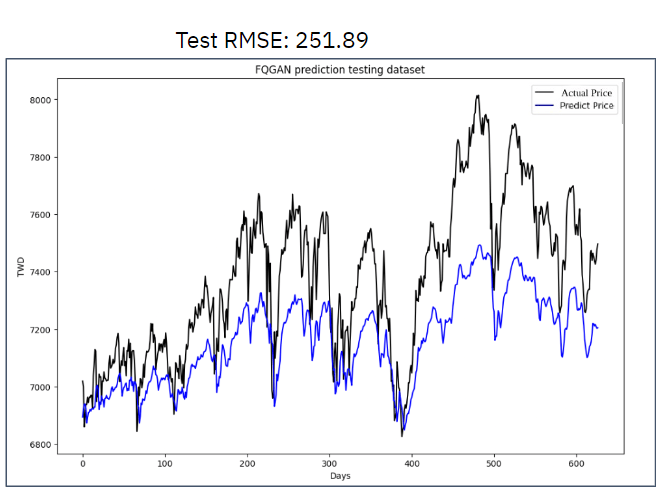}
   \end{minipage}
   \centering
   \caption{Full Quantum GAN. Train (up) and test (down).}
   \label{fig:full_qgan_trainNtest}
\end{figure}

\textbf{Performance Metrics:} Trained for just 5 epochs on the AWS SV1 quantum simulator with a learning rate of 0.016, the FQGAN model showed considerable promise despite its short training duration. The model achieved a training RMSE of 571.36 and a test RMSE of 251.89, suggesting a competitive performance, particularly when considering the limited quantum resources and training epochs.
Fig. \ref{fig:full_qgan_trainNtest}.

\subsubsection{Results across Varying Window Sizes}

\begin{figure}
\centering
    \includegraphics[width=0.9\linewidth]{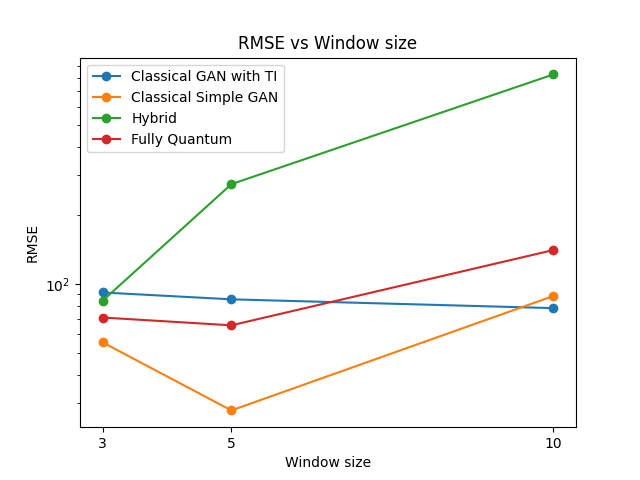}
    \caption{RMSE for the models across varying past window sizes}
    \label{fig:results_on_window_size}
\end{figure}

We also explored the effect of varying past window sizes on the performance of the four models. The results indicate that the Hybrid Quantum GAN exhibits a decline in performance as the window size increases, while the FQGAN's RMSE improves up to a window size of 5, before deteriorating at a window size of 10. Fig. \ref{fig:results_on_window_size}

\subsubsection{Results on AWS using SV1 Simulator}
While running the experiments on the SV1 quantum simulator, we faced challenges due to the unavailability of real hardware devices and extended queue times. Each epoch took two hours, and with a minimum of 5 epochs required to obtain significant results, the total simulation time for each model stretched beyond 10 hours. Due to these limitations, we were unable to perform tests on real hardware.

\begin{figure}[h!]
\centering
   \begin{minipage}{0.45\textwidth}
     \centering
     \includegraphics[width=\linewidth]{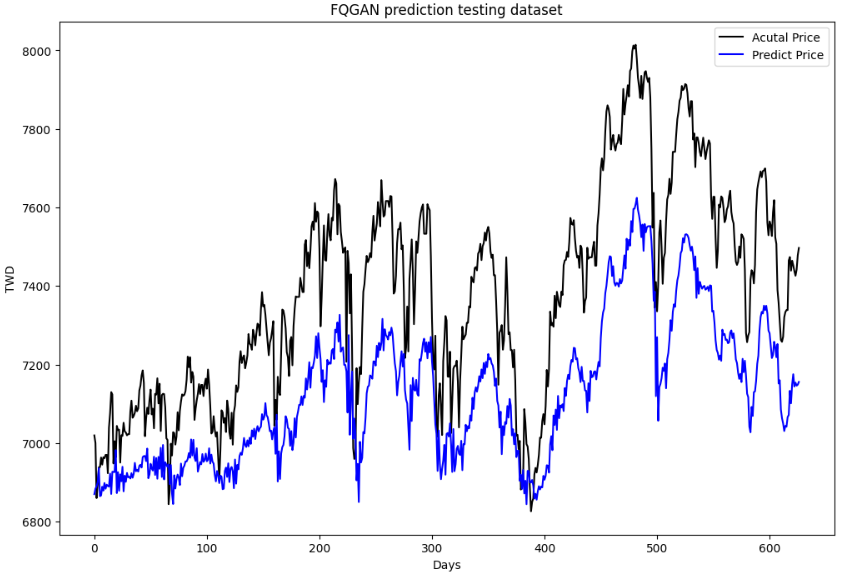}
     \centering
     \includegraphics[width=\linewidth]{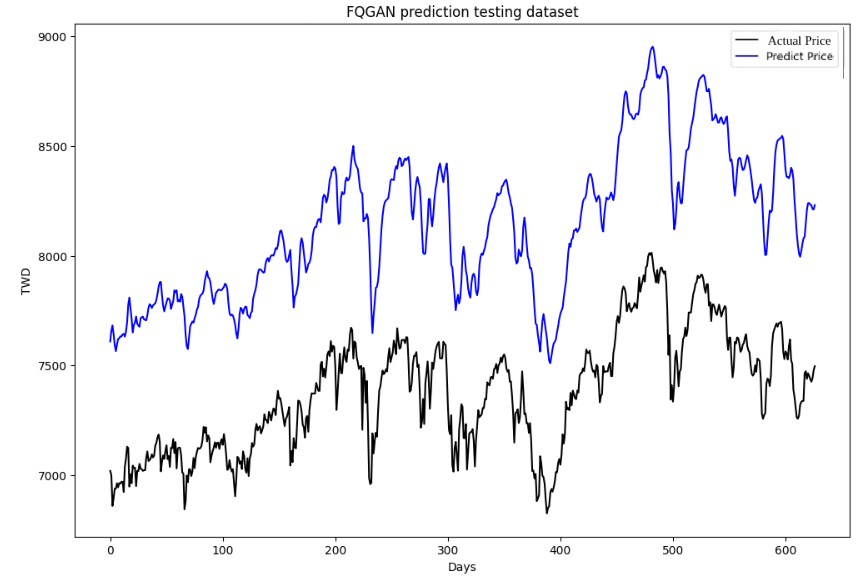}
   \end{minipage}
   \centering
   \caption{From AWS SV1, 3 epochs (up) and 2 epochs (down).}
   \label{fig:aws_sv1_results}
\end{figure}

The Classical GAN achieved a testing RMSE of 91.71, which serves as the standard for comparing the hybrid and fully quantum models. Notably, the Fully Quantum GAN performed poorly in comparison, which was expected due to its nascent stage of development. The hyperparameters, particularly the learning rate, played a significant role in model performance, with lower learning rates improving test accuracy. Fig. \ref{fig:aws_sv1_results}

\subsection{Predicting a larger window size for the future}

We expanded our experiment to predict larger future window sizes using Simple GAN and Fully Quantum GAN. The results show that FQGAN performs better with smaller datasets, even with larger window sizes, indicating the model’s ability to generalize effectively with fewer data points.

From Fig~\ref{fig:metrics}, we can observe the values of RMSE, MAE and R2 scores. The metric values are consistently lower for the FQGAN relative to the simple GAN. As we increase the past and future window sizes, the size of the dataset decreases. For example, with a window size of (4,2) (past window, future window), there are nearly 1500 data points; however, with a window size of (32, 16), it reduces to 194. From the figure, it is evident that the FQGAN can perform better than Classical GAN even with fewer data points to train on. 

\subsubsection{Quantum Circuit Resource Usage}

\begin{figure*}[h!]
    \includegraphics[width= 1.0 \linewidth]{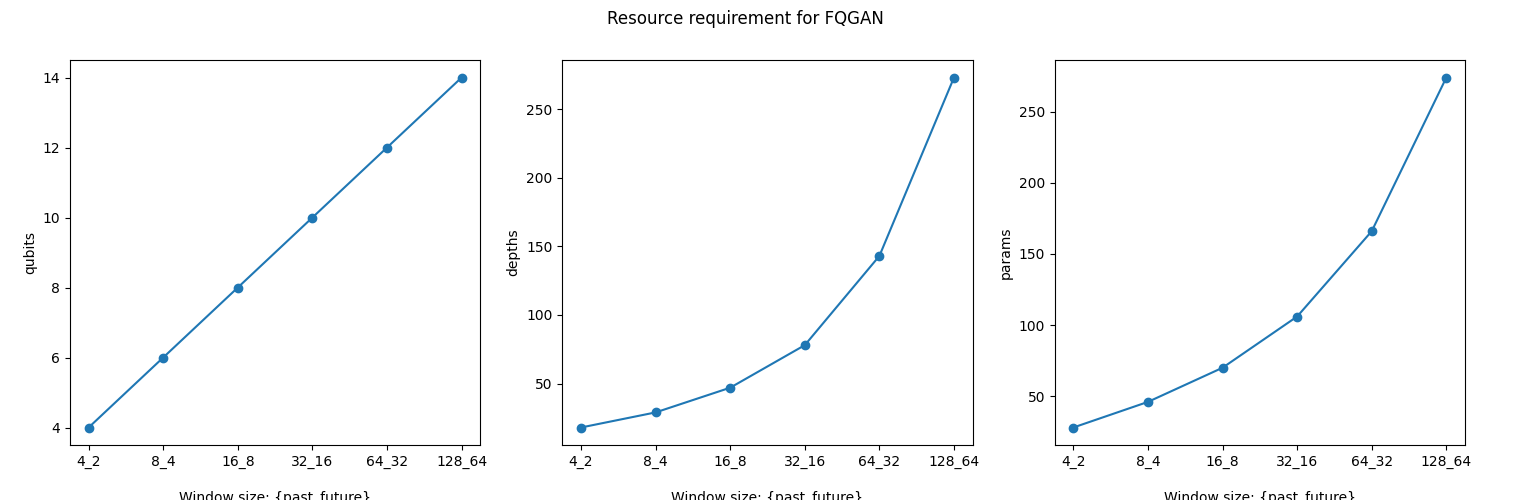}
    \caption{Number of qubits (left), circuit depth (middle) and number of trainable parameters (right) for fully quantum GAN. The x-axis has the different window sizes.}
    \label{fig:um_qubits_VS_ckt_depth}
\end{figure*}

From Eq.~\ref{eq:qubit}, we can calculate the number of qubits required by the FQGAN circuit. The depth of the circuit scales exponentially because we use Amplitude Embedding for encoding the data. One way to reduce the depth of the circuit is to use efficient data embedding schemes. \cite{31} proposes a parallel amplitude embedding method which can reduce the depth of the amplitude embedding circuit by 25\% for more than 10 qubits \cite{32}. Fig. \ref{fig:um_qubits_VS_ckt_depth}

\section{Improvement}
The potential of quantum computing in the financial services industry is both substantial and transformative. Quantum algorithms, particularly those designed for predictive modeling, can significantly enhance the accuracy and efficiency of stock price forecasting. By processing vast amounts of market data, including historical prices, trading volumes, and economic indicators, Quantum Machine Learning (QML) models possess the ability to identify complex patterns that may remain elusive to classical machine learning algorithms.

In this context, our proposed solution—employing Quantum Generative Adversarial Networks (QGANs) for stock price prediction—represents a significant advancement in the field. Through the integration of quantum computing, we have achieved a quantum advantage over traditional methods, particularly in terms of the ability to predict stock prices with smaller datasets. The novel approach of utilizing both a quantum generator and a quantum discriminator has proven instrumental in enhancing model performance, yielding superior predictive outcomes compared to classical counterparts. This breakthrough has far-reaching implications for fund managers, providing them with highly accurate price predictions that can inform and optimize investment decisions. By enabling more precise stock price forecasts, fund managers are better equipped to refine their investment strategies, mitigate risks, and ultimately maximize returns.

Our work emphasizes the use of quantum computing to refine prediction accuracy, empowering fund managers to make data-driven, informed decisions and optimize portfolio performance across various time horizons. By integrating quantum-powered price predictions for both market indices and individual stocks, we enable investors to construct portfolios that not only minimize potential risks but also generate superior returns. Additionally, our contributions to individual stock analysis have significantly enhanced fund management capabilities, allowing financial institutions to provide clients with more effective and tailored investment strategies.

Beyond improving decision-making processes, our approach also strengthens portfolio performance and risk management strategies. By providing accurate stock price forecasts and enabling insightful portfolio analysis, we facilitate a more strategic investment approach. Leveraging state-of-the-art quantum computing technologies, our methodology empowers securities departments to adopt a forward-looking approach to trading decisions, ensuring optimal returns for clients and enhancing overall financial outcomes. \cite{27, 30}

\begin{figure*}
    \includegraphics[width=1.0\linewidth]
    {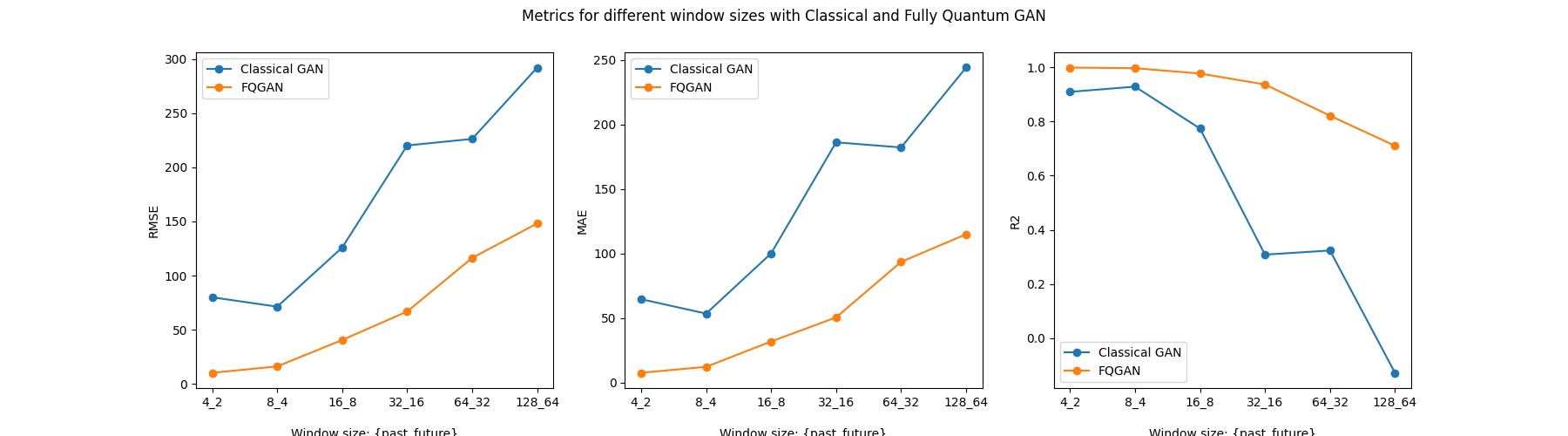}
    \caption{RMSE (left), MAE (middle) and R2 (right) for classical simple and fully quantum GAN. The x-axis has the different window sizes.}
    \label{fig:metrics}
\end{figure*}

\section{Conclusion} 

This research provides a comprehensive exploration of the application of Quantum Generative Adversarial Networks (QGANs) for stock index price forecasting, making a significant contribution to the emerging field of Quantum Finance. Through a comparative analysis of classical GANs, hybrid quantum-classical GANs, and fully quantum GANs, we highlight the superior performance of quantum-based models relative to conventional forecasting techniques.

Empirical results consistently demonstrate that quantum-enhanced models, particularly fully Quantum GANs, outperform classical methodologies such as ARIMA, LSTM, and traditional GANs in terms of both accuracy and computational efficiency. The observed advantage is largely attributed to the unique quantum properties of superposition and entanglement, which enable these models to process and analyze complex, high-dimensional data more effectively than their classical counterparts. \cite{14, 15, 16}

A key innovation introduced in this study is the development of the Invertible Fully Quantum GAN (FQGAN), which addresses the normalization challenges inherent in quantum models. This advancement enhances the stability and reliability of quantum-based predictions, offering promising results in stock price forecasting and underscoring the practical viability of quantum computing techniques in financial market predictions.

The implications of these findings are far-reaching, particularly for industries within the financial sector. Quantum computing presents opportunities to revolutionize predictive analytics, optimize portfolio management, and refine investment strategies. As quantum computing continues to evolve, these advancements are expected to drive significant improvements in financial modeling and decision-making processes.

In conclusion, this research not only validates the potential of QGANs in stock price prediction but also establishes a robust foundation for further exploration in Quantum Finance. As quantum technologies mature, we anticipate continued progress in the refinement of predictive models, with applications extending beyond stock price forecasting to a broader range of financial domains. The advent of quantum computing in the financial industry represents a paradigm shift, offering immense opportunities for future research and innovation. \cite{31}

\section*{Acknowledgement}

The authors would like to recognize AWS Braket, as all simulations were performed on SV1 simulators. All the authors would also like to thank Prof. Greg Byrd and Prof. Kazuki Ikeda for the meaningful discussions and comments.

\end{document}